\documentclass[runningheads]{llncs}
\usepackage{textcomp}
\usepackage[T1]{fontenc}
\usepackage{etoolbox}

\usepackage{comment}

\usepackage{float}

\usepackage{textcomp}

\usepackage{todonotes}
\usepackage{multicol}
\usepackage{algpseudocode}
\usepackage{amsmath, amssymb}
\usepackage{graphicx}
\usepackage{subcaption} 
\usepackage{paralist}
\usepackage{lmodern} 
\usepackage{tabularx}
\usepackage{moresize}
\usepackage{rotating}
\usepackage{caption}
\usepackage{url}
\usepackage{float}
\usepackage{lmodern}

\usepackage{multirow}
\usepackage{booktabs}
\usepackage{makecell}
\usepackage{multirow}
\usepackage{setspace}
\usepackage{bm}
\usepackage{hyperref} 
\usepackage{placeins} 
\FloatBarrier 

\begin{document}
%
\title{Exploring Model Quantization in GenAI-based Image Inpainting and Detection of Arable Plants}

\titlerunning{Model Quantization for GenAI Image Inpainting \& Plant Detection}

%
\author{Sourav Modak~\orcidID{0009-0006-7759-0864} \and
Ahmet O\u{g}uz Salt{\i}k~\orcidID{0000-0002-8873-1638} \and
Anthony Stein~\orcidID{0000-0002-1808-9758} }
\authorrunning{Modak, Saltik and Stein}
%
\institute{University of Hohenheim, Dept. Artificial Intelligence in Agricultural Engineering \& Computational Science Hub, 70599 Stuttgart, Germany
\email{\{s.modak,ahmet.saltik,anthony.stein\}@uni-hohenheim.de}\\}
\maketitle              
\renewcommand\thefootnote{}
\footnotetext{\textit{Preprint submitted to the Special Track on Organic Computing at the Architecture of Computing Systems (ARCS) 2025 conference}.}
\begin{abstract}
Deep learning-based weed control systems often suffer from limited training data diversity and constrained on-board computation, impacting their real-world performance. To overcome these challenges, we propose a framework that leverages Stable Diffusion-based inpainting to augment training data progressively in 10\% increments—up to an additional 200\%, thus enhancing both the volume and diversity of samples. Our approach is evaluated on two state-of-the-art object detection models, YOLO11(l) and RT-DETR(l), using the mAP50 metric to assess detection performance. We explore quantization strategies (FP16 and INT8) for both the generative inpainting and detection models to strike a balance between inference speed and accuracy. Deployment of the downstream models on the Jetson Orin Nano demonstrates the practical viability of our framework in resource-constrained environments, ultimately improving detection accuracy and computational efficiency in intelligent weed management systems.

\end{abstract}

\section{Introduction}
Automated weed control systems, such as robots and smart sprayers, play a vital role in enhancing weeding efficiency, improving crop health, and increasing yields. In contrast, manual weed management is both costly and time-consuming, typically eliminating only 65--85\% of weeds~\cite{hu2024real}. However, achieving viable real-world agricultural performance requires a high-quality, diverse, and balanced dataset. Traditional image augmentation techniques (e.g., image manipulation, geometric transformations, copy-paste, and mixup) often fall short in generating the necessary diversity, as they depend heavily on the input data. In contrast, synthetic images produced by Generative Adversarial Networks (GANs)~\cite{goodfellow2014generative} and Diffusion Models~\cite{dhariwal2021diffusion} provide highly diverse and realistic representations while preserving natural variations~\cite{10.1007/978-3-031-66146-4_8}~\cite{tudosiu2024realistic}. In our previous study, synthetic data augmentation based on the Stable Diffusion Model~\cite{Rombach_2022_CVPR} has been found to potentially improve the performance of downstream models on a sugar beet weed detection task~\cite{modak2024enhancing}. Notably, replacing only 10\% of real-world images with synthetic ones was observed to be sufficient to enhance training efficiency~\cite{modak2024generative} in the investigated case of weed detection. Yet, this approach was limited by its sole reliance on synthetic data generation, without the ability to manipulate existing images. To overcome this limitation, we extend our methodology by incorporating inpainting techniques~\cite{TAO2024111082}. This extension not only enables precise spatial manipulation, weed size control, and class imbalance reduction within existing images but also seamlessly integrates synthetic augmentation into real-world datasets. 
In this paper, we evaluate our enhanced approach by progressively incorporating synthetic images during the training of state-of-the-art models such as YOLO11(l) and RT-DETR(l). While these techniques achieve substantial performance gains in contrast to nano and small models, deploying real-time capable deep learning models in practical setups such as agricultural machines for weed detection without sacrificing detection accuracy is still challenging, since larger models with higher predictive capability demand considerable higher computational power~\cite{rai2023applications}. 
To address these challenges, we investigate model quantization techniques purposefully reducing the numerical precision of neural network weights and activations  (e.g., \textit{FP16} and \textit{INT8} precision) within a generative AI-based weed detection pipeline introduced earlier~\cite{10.1007/978-3-031-66146-4_8} which integrates a newly developed image inpainting method and downstream weed detection models. 
To explore the performance impacts on resource-constrained edge devices as presumably used in future intelligent weed control systems, we deploy our downstream detection models on an NVIDIA Jetson Orin Nano\footnote{\url{https://developer.nvidia.com/embedded/learn/get-started-jetson-orin-nano-devkit} (accessed on February 14, 2025)}.
The contributions of this paper are two-fold and can be summarized as providing initial answers to the following two resarch questions:

\begin{enumerate}
    \item What is the effect of post-training quantization on GenAI inpainting and downstream weed detection accuracy?
    \item To what extent does post-training quantization enhance inference efficiency while balancing computational trade-offs in the Stable Diffusion and downstream models?
\end{enumerate}


\section{Background}
\label{sec:rel}

\subsection{Related Work}

Recent advances in generative AI-based image generation have paved the way for a variety of data augmentation techniques that significantly enhance task performance in agricultural applications. Techniques such as text-prompt-based image generation, image inpainting, and image-to-image translation have been widely adopted, with diffusion models playing a pivotal role. For instance, Deng et al.~\cite{deng2025weed} demonstrated a 1.5\% improvement in mAP50-95 scores through enhanced image augmentation. Similarly, Modak \& Stein~\cite{modak2024enhancing} showed that applying Stable Diffusion to crop-weed images improved YOLO model performance more effectively than traditional augmentation methods. Moreover, domain adaptive data augmentation using diffusion models has been reported to benefit vineyard shoot detection~\cite{hirahara2025d4}. In parallel, model quantization techniques have emerged as a promising strategy to reduce computational demands while preserving inference speed and detection accuracy. Notably, INT8 quantization offers an effective compromise by optimizing computational requirements without significant sacrifices in performance~\cite{herterich4862267accelerating}. 

\subsection{Diffusion Model}

Diffusion models, a subclass of generative models, operate by corrupting training data with Gaussian noise (known as \textit{forward diffusion}) and learning to recover the original information through step-by-step denoising (known as \textit{reverse diffusion}). Given an initial data distribution \( x_0 \sim q(x) \), the forward process adds Gaussian noise via a Markovian transition:
\vspace{-5pt}
\begin{equation}
    q(x_t | x_{t-1}) = \mathcal{N}(x_t ; \sqrt{1 - \beta_t} x_{t-1}, \beta_t I),
\end{equation}

where \( \beta_t \) is a variance schedule controlling noise addition. Reparameterizing, the noisy data at timestep \( t \) is:
\vspace{-5pt}
\begin{equation}
    x_t = \sqrt{\alpha_t} x_0 + \sqrt{1 - \alpha_t} \epsilon, \quad \epsilon \sim \mathcal{N}(0, I),
\end{equation}

where \( \alpha_t = \prod_{s=1}^{t} (1 - \beta_s) \) represents the cumulative noise factor. The model aims to reverse this process by learning the posterior:
\vspace{-5pt}
\begin{equation}
    p_\theta(x_{t-1} | x_t) = \mathcal{N}(x_{t-1}; \mu_\theta(x_t, t), \Sigma_\theta(x_t, t)),
\end{equation}

where the mean function is parameterized as:
\vspace{-5pt}
\begin{equation}
    \mu_\theta(x_t, t) = \frac{1}{\sqrt{1 - \beta_t}} \left( x_t - \frac{\beta_t}{\sqrt{1 - \alpha_t}} \epsilon_\theta(x_t, t) \right).
\end{equation}

A U-Net with sinusoidal time embeddings approximates \( \epsilon_\theta(x_t, t) \). The model is trained by minimizing the MSE loss:
\vspace{-5pt}
\begin{equation}
    L(\theta) = \mathbb{E}_{x_0, t, \epsilon} \left[ \|\epsilon - \epsilon_\theta(x_t, t)\|^2 \right].
\end{equation}

 The Stable Diffusion Model, based on the Latent Diffusion Model (LDM), enables conditional text-to-image generation using CLIP text embeddings. Performing the diffusion process in latent space instead of pixel space, significantly boosts efficiency. A trained encoder maps high-resolution images to a lower-dimensional latent representation, which is then reconstructed by a decoder. Stable Diffusion's conditioning mechanism allows for text prompts, input masks, and layouts, supporting tasks including text-to-image generation, inpainting, and super-resolution. To enhance computational efficiency during inference, various sampling schedulers such as; Denoising Diffusion Implicit Models (DDIM) and the Euler Ancestral sampler are used~\cite{Rombach_2022_CVPR}. 
\subsection{Object Detection Models}
\label{sub:obj_dect}
\paragraph{\textbf{YOLO11}}
Evolving from the YOLO family, YOLO11 is a state-of-the-art real-time object detector model. It enhances object detection speed and accuracy with key architectural innovations. The model uses initial convolutional layers for downsampling, followed by the C3k2 block for improved computational efficiency. Features including Spatial Pyramid Pooling-Fast (SPPF) and a new Cross Stage Partial with Spatial Attention module enhance focus on salient regions, improving small object detection. In the neck, features from various scales are fused, while the head refines them through additional C3k2 and CBS layers, ultimately producing bounding boxes, objectness scores, and class predictions~\cite{ghosh2024yolo11}. 

\paragraph{\textbf{RT-DETR}} 
In contrast to convolution-based architectures, RT-DETR uses a transformer-based backbone to analyze the entire image, allowing it to capture global context for detecting complex scenes and small objects. It eliminates the need for non-maximum suppression (NMS) by using a one-to-one matching strategy with the Hungarian algorithm, resulting in unique predictions and faster detection. The architecture features a hybrid encoder that combines Attention-based Intra-scale Feature Interaction (AIFI) with CNN-based Cross-scale Feature Fusion (CCFF) for effective multi-scale feature extraction. Additionally, a query selection mechanism improves the quality of initial queries, enhancing overall detection accuracy~\cite{zhao2024detrs}.

\paragraph{\textbf{Object Detection Metrics}}
The efficiency of object detection models is evaluated using metrics such as \textit{precision, recall, mAP50,} and \textit{mAP50--95}. \textit{Precision} measures the proportion of correctly identified positives among all predicted positives, while \textit{recall} measures the proportion of actual positives that are correctly detected. Beyond classification, localization accuracy is crucial in detection tasks. \textit{Intersection over Union (IoU)} quantifies the overlap between predicted and ground-truth bounding boxes, with detections considered correct if $\text{IoU} \geq t$ (\textit{t} is a predefined threshold). \textit{Precision-recall (PR) curves} illustrate the trade-off between precision and recall, with the \textit{area under the curve (AUC)} indicating overall performance. \textit{Average Precision (AP)} computes AUC for a single class, while \textit{Mean Average Precision (mAP)} averages AP across classes. \textit{mAP50} is evaluated at $\text{IoU} = 0.50$, whereas \textit{mAP50--95} averages \textit{mAP} over thresholds from 0.50 to 0.95. 

\subsection{Quantization}
\label{sub:quan}
Deep learning models typically use 32-bit floating-point (FP32) arithmetic, while lower-precision formats (FP16 \& INT8) are often employed for efficiency.

\textit{\textbf{FP32 Representation}}
FP32 (single precision) uses 32 bits -- 1 bit for the \textit{sign}, 8 bits for the \textit{exponent} (bias 127), and 23 bits for the \textit{mantissa}. A floating-point number \( x \) in FP32 is:
\vspace{-8pt}
\begin{equation}
x = (-1)^s \times 2^{e-127} \times \left(1 + \frac{m}{2^{23}}\right),
\end{equation}
where \( s \) is the sign bit, \( e \) is the exponent, and \( m \) is the mantissa.

\textit{\textbf{FP16 Representation}}  
FP16 (half precision) uses 16 bits -- 1 bit for the \textit{sign}, 5 bits for the \textit{exponent} (bias 15), and 10 bits for the \textit{mantissa}. The FP16 representation of a number \( x \) is:
\vspace{-8pt}
\begin{equation}
x = (-1)^s \times 2^{e-15} \times \left(1 + \frac{m}{2^{10}}\right),
\end{equation}
where \( s \) is the 1-bit sign, \( e \) is the 5-bit exponent, and \( m \) is the 10-bit mantissa.

\textit{\textbf{INT8 Quantization}}  
INT8 quantization maps a full-precision weight \( w \) to an 8-bit integer \( q \). In a signed INT8 representation (\([-128,127]\)), the conversion follows:
\vspace{-5pt}
\begin{equation}
q = \operatorname{clip}\left(\operatorname{round}\left(\frac{w}{s}\right) + z,\, -128,\, 127\right),
\end{equation}
where \( s \) is the scaling factor, \( z \) is the zero-point, and \(\operatorname{clip}\) ensures \( q \) remains in range:
\vspace{-5pt}
\begin{equation}
\operatorname{clip}(x, a, b) = \max(a, \min(b, x)).
\end{equation}

The original weight is approximately recovered as:
\vspace{-5pt}
\begin{equation}
\hat{w} = s \cdot (q - z).
\end{equation}

These quantization methods apply in both post-training quantization and Quantization-Aware Training (QAT), where the loss accounts for quantization error:
\vspace{-5pt}
\begin{equation}
L_{\text{total}} = L\left(\hat{w}\right) + \lambda \, \|w - \hat{w}\|^2,
\end{equation}
with \(\lambda\) balancing task loss and quantization error. While FP32 and FP16 explicitly represent numbers using separate fields for sign, exponent, and mantissa, INT8 uses an integer representation with a scaling factor and zero-point. These quantization techniques trade numerical precision for efficiency, critical for real-time object detection models~\cite{IEEE754}\cite{Krishnamoorthi2018}.

\section{Material and Methods}
\label{sec:mat}

\begin{figure}[h!]
    \centering
    \includegraphics[width=1.0\linewidth]{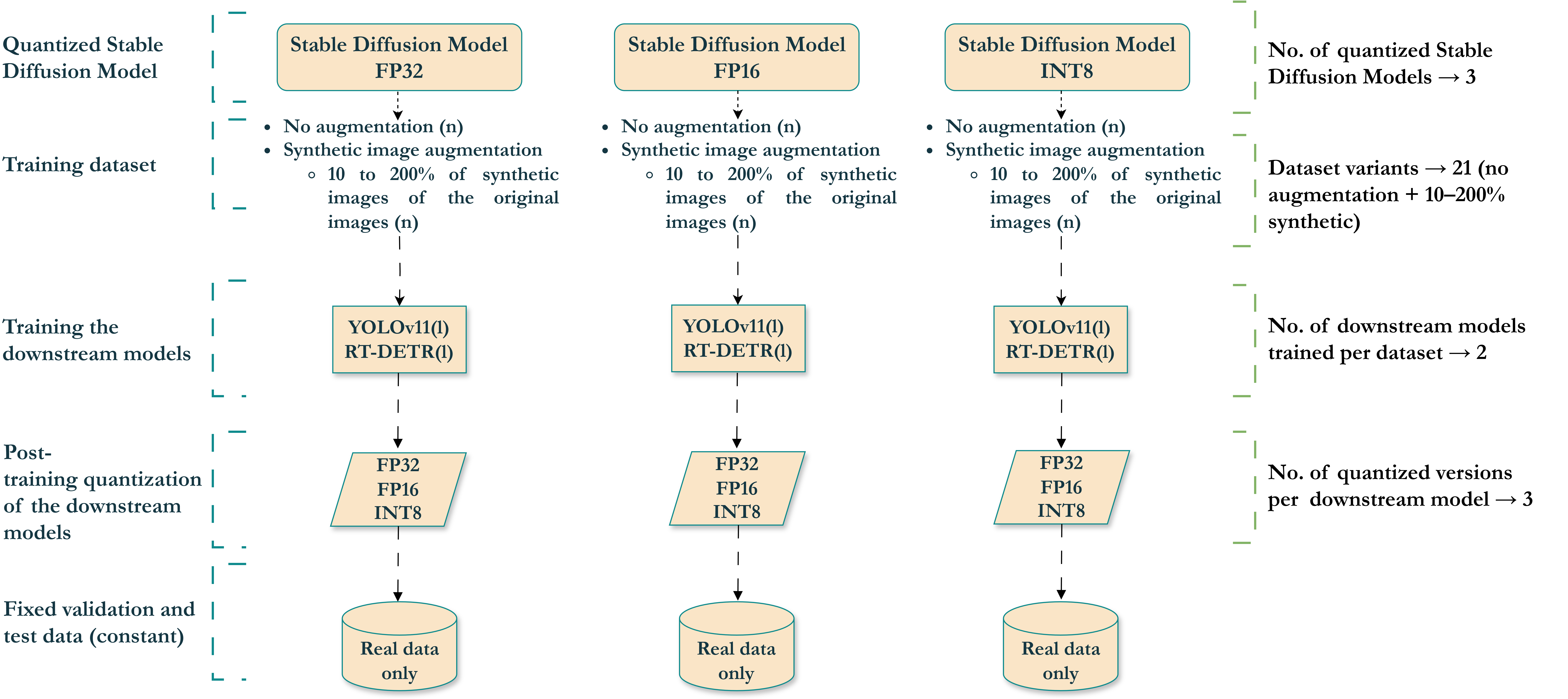}
    \caption{Workflow of the quantization process and training of downstream models.
 Image generation using Stable Diffusion models at different precision levels. Synthetic image augmentation (10\%–200\% of the original dataset). Two downstream models (YOLO11(l) and RT-DETR(l)) are trained per dataset and later quantized and evaluated on a fixed set of real validation and test data.}
    \label{fig:quant}
\end{figure}

Accurate weed identification is vital for intelligent agricultural weed management systems and depends on high-quality data. Our method addresses real-world data challenges, including the under-representation of certain weed classes due to factors, namely seasonal changes, weather, and seed availability. To address these challenges, we utilize an image augmentation strategy derived from our previous data augmentation work~\cite{10.1007/978-3-031-66146-4_8}, which employs an inpainting technique (refer to Fig.~\ref{fig:pipe}). This strategy augments the dataset by generating images of underrepresented weed classes, enhancing its diversity and volume. YOLO variants offer a good balance of accuracy and speed, while RT-DETR achieves higher accuracy with slower inference \cite{saltik2024comparative} \cite{allmendinger2025assessing}. Our previous work \cite{allmendinger2025assessing} shows that YOLO(l) and RT-DETR(l) match their YOLO(x) and RT-DETR(x) counterparts in accuracy while improving increasing efficiency. Thus, we validated our augmented images by fine-tuning YOLO11(l) and RT-DETR(l) on original and synthetic datasets.   The proportion of augmented data varied, ranging from 10\% to 200\% of the original dataset size, with a 10\% increment. However, both the inpainting image generation and subsequent weed detection are computationally intensive. We addressed this by applying post-training quantization (see~\ref{sub:quan}) to speed up inference and lower resource demands (see Fig.~\ref{fig:quant}). For training Stable Diffusion, YOLO11(l), and RT-DETR(l), we used an NVIDIA A100-SXM4-40GB GPU alongside an AMD EPYC 75F3 32-core processor with 12 GB of memory. Our quantized, fine-tuned downstream models were then deployed on an NVIDIA Jetson Nano 8GB variant, featuring an 8-core ARM Cortex-A78AE CPU and a 1024-core Ampere GPU with 8 GB of unified memory. 

\subsection{Dataset}
\label{subsec:data}

\begin{figure}[h!]
    \centering
    \begin{subfigure}{0.35\textwidth}
        \centering
        \includegraphics[width=\textwidth]{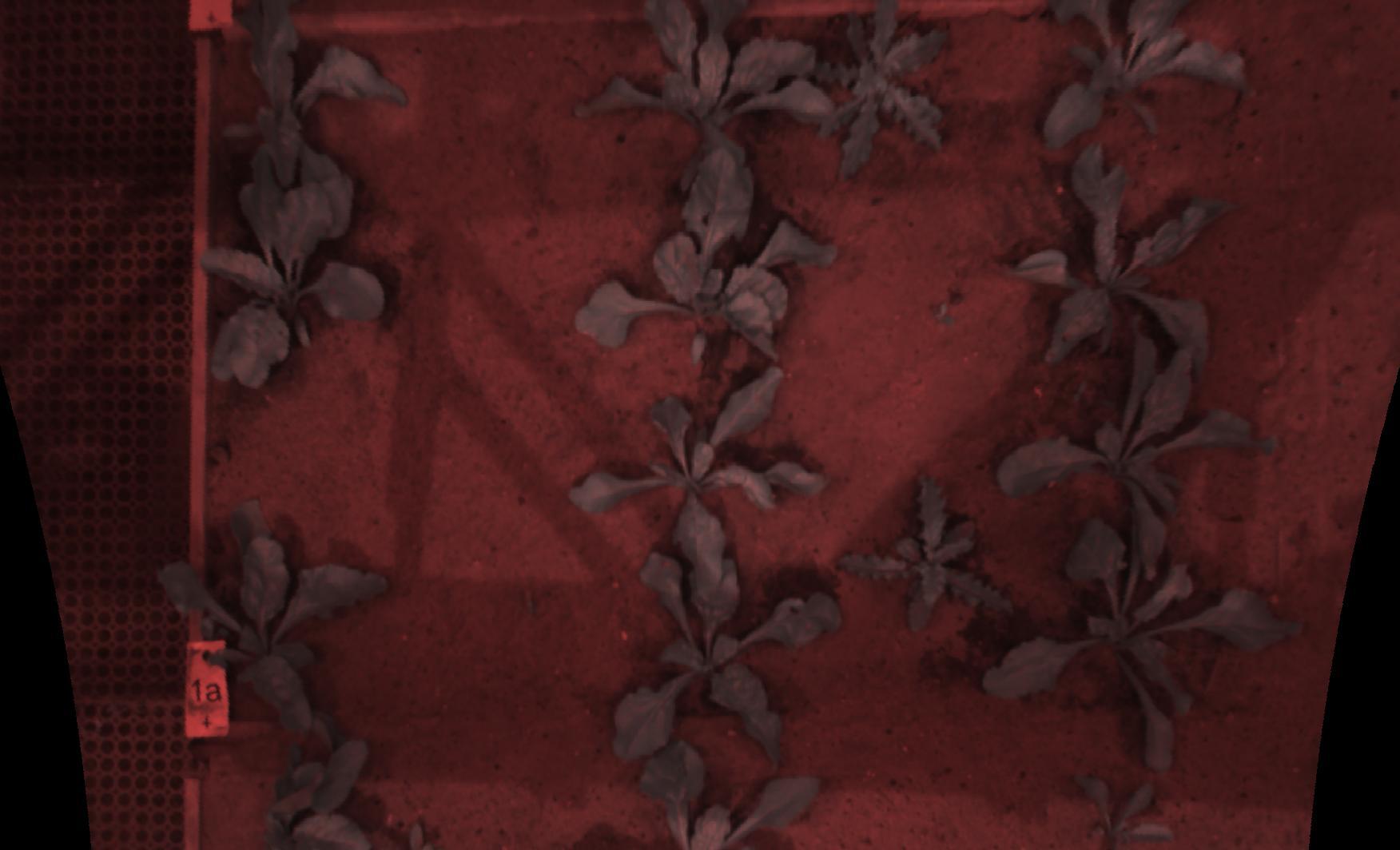}
    \end{subfigure}
    \hspace{0.5cm} 
    \begin{subfigure}{0.35\textwidth}
        \centering
        \includegraphics[width=\textwidth]{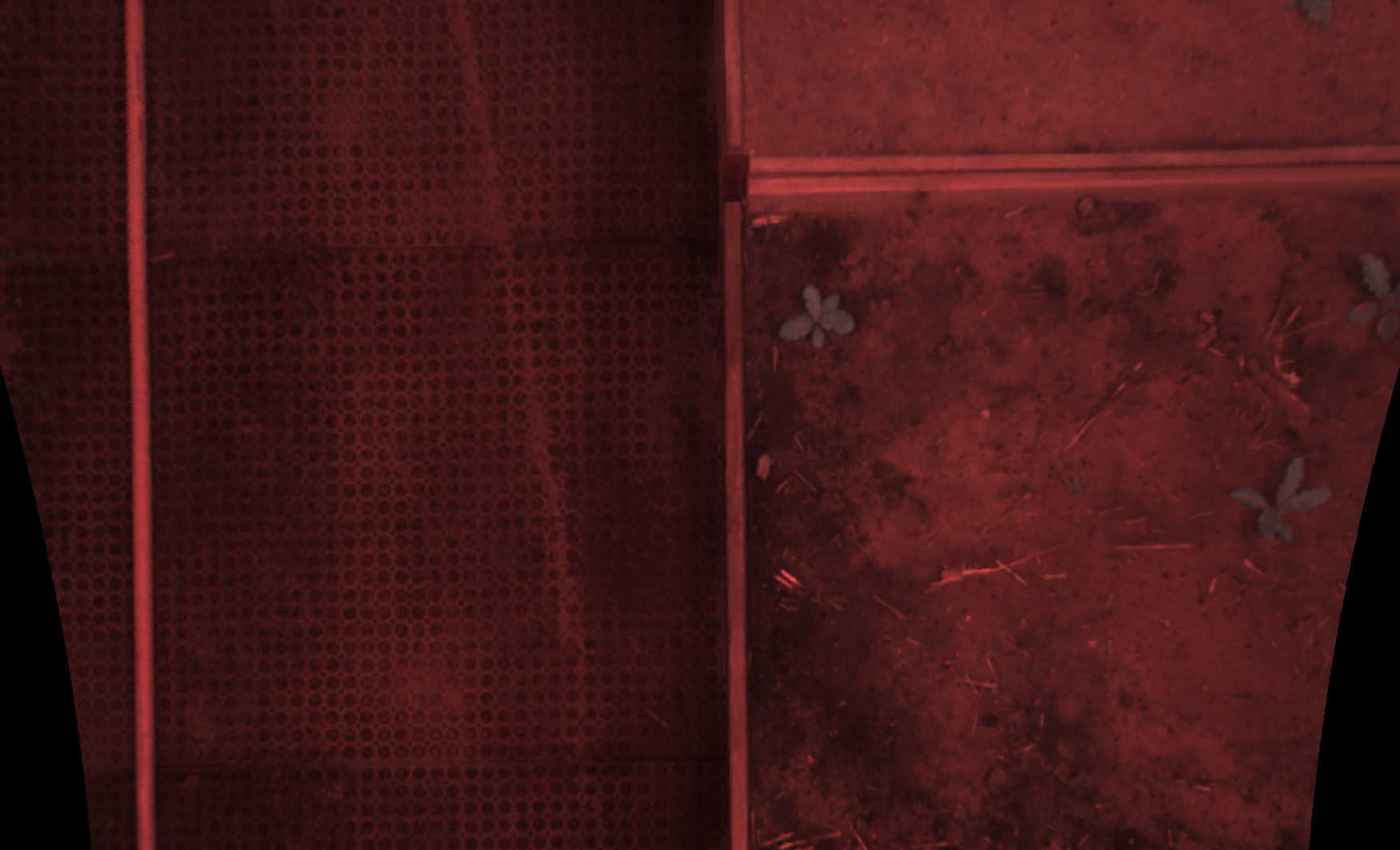}
    \end{subfigure}
    \caption{A representative sample of pseudo-RGB images highlighting sugar beet crops distributed with various weed species on the euro-pallets.}
    \label{fig:RGB}
\end{figure}

The dataset was collected from a test site using a \textit{field camera unit (FCU)} mounted on a smart sprayer attached to a tractor moving at \(1.5 \mathrm{m}/\mathrm{s}\). The imaging system featured a 6 mm \textit{effective focal length (EFL)} and 2.3 MP RGB sensors, with a dual-band lens filter for near-infrared (NIR) and red wavelengths. Post-processing included projection correction and pseudo-RGB image generation from NIR and red wavelengths. Images were captured from 1.1 m above ground at a 25-degree tilt. The dataset, comprising 2074 images, includes \textit{Sugar beet} as the main crop and four weed types: \textit{Cirsium, Convolvulus, Fallopia, and Echinochloa} under diverse soil condition. These images were precisely annotated by field experts with a background in agronomic studies for object detection purposes. Each image has a resolution of $ 1752\times1064$ pixels (see Fig.~\ref{fig:RGB}).

\subsection{Experimental Settings}
\paragraph{\textbf{Pipeline Architechture}}

\begin{figure}[h!]
    \centering
    \includegraphics[width=1.0\linewidth]{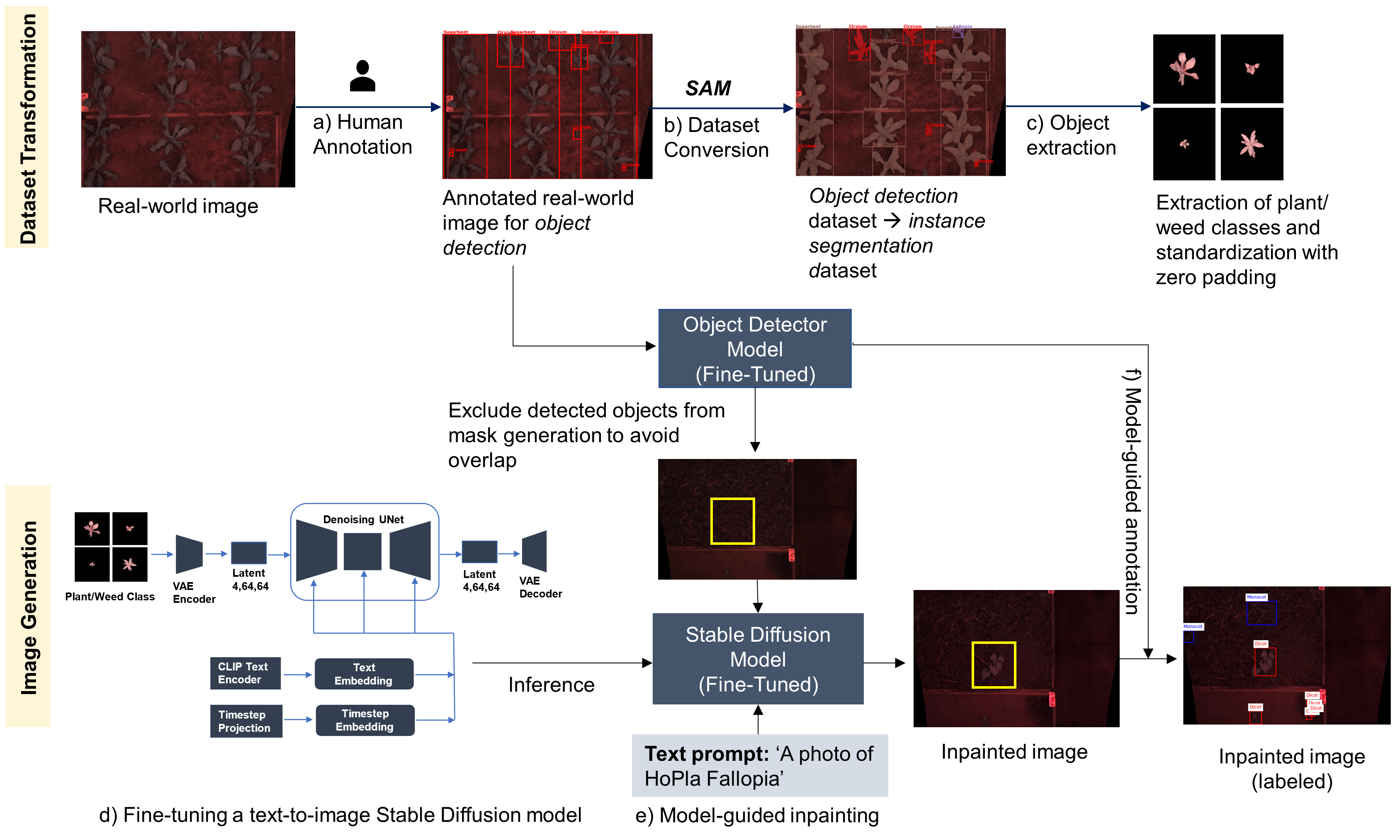}
    \caption{Overview of the proposed inpainting pipeline architecture. In (a), real-world images are manually annotated. In (b), the SAM converts user-specified bounding boxes into precise polygon masks, which facilitate object extraction in (c). In (d), a Stable Diffusion model is fine-tuned using the extracted plants and weeds. A novel inpainting method is then applied in (e), using a simple prompt (e.g., “A photo of HoPla Fallopia”) along with dynamically generated inpainting masks to indicate where new plant or weed elements should be inserted, while a fine-tuned object detector prevents overlap with existing objects. Finally, in (f), the same object detector is used to label the inpainted images.}
    \label{fig:pipe}
\end{figure}

The proposed pipeline comprises two main components: dataset transformation and image generation (cf. Fig.~\ref{fig:pipe}). The dataset transformation stage is derived from our preceding study~\cite{10.1007/978-3-031-66146-4_8}. Originally intended for object detection purposes, the dataset is transformed into a zero-shot setting using the Segment Anything Model (SAM)~\cite{kirillov2023segany}, specifically the \textit{SAM ViT-H} variant, which converts bounding box annotations into polygonal masks. The plant and weed shapes are then isolated and the images are standardized by zero-padding. During the image generation phase, the Stable Diffusion Model v1.5 is fine-tuned on the extracted plant and weed classes using the diffusers~\cite{von-platen-etal-2022-diffusers} library, employing a specific technique denoted as \textit{multi-subject Dreambooth}. Table~\ref{tab:hyp_SD} outlines the hyperparameters utilized during the training process. 
Given constraints on GPU memory, a batch size of 1 was used, along with gradient checkpointing. For stable and smooth convergence, learning rate \(5\times10^{-6}\) was applied with a cosine learning rate scheduler. Moreover, to enhance memory usage and computational efficiency, a dynamic quantization method, also known as FP16 mixed precision training, was implemented. 
Additionally, a text encoder was trained with a unique identifier, namely, HoPla~\footnote{\url{https://www.photonikforschung.de/projekte/sensorik-und-analytik/projekt/hopla.html} (accessed on February 28, 2025)}, alongside subject classes, including \textit{Sugar beet}, \textit{Cirsium}, \textit{Fallopia}, and \textit{Convoluvulus}. 

\begin{table}[h!]
\scriptsize
\centering
\renewcommand{\arraystretch}{0.85}
\caption{Hyperparameter configuration for Stable Diffusion Model training using the diffusers library.}
\begin{tabularx}{\linewidth}{@{\extracolsep{\fill}} lX | lX }
\toprule
\textbf{Hyperparameter} & \textbf{Value} & \textbf{Hyperparameter} & \textbf{Value} \\
\midrule
Image resolution        & 512                     & Epoch                   & 2 \\
Batch size              & 1                       & Gradient checkpointing  & True \\
Learning rate           & \(5\times10^{-6}\)      & Learning rate scheduler & cosine \\
Maximum training steps  & 60000                   & Mixed precision         & FP16 \\
\bottomrule
\end{tabularx}
\label{tab:hyp_SD}
\end{table}

\begin{figure}[h!]
    \centering
   
        \centering
        \includegraphics[width=0.9\textwidth]{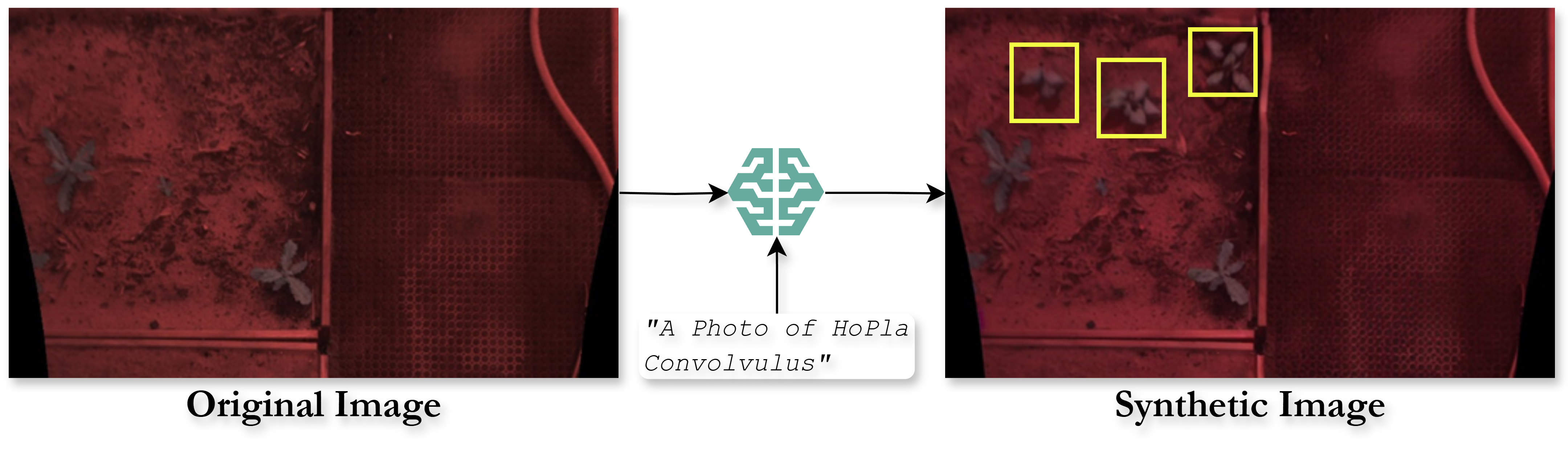}
    \caption{ A visual representation of image inpainting.  The left side shows the original image, while the right side displays the synthetic image generated using a text prompt -- \textit{`A photo of HoPla Convolvulus'}, where the highlighted regions indicate the inpainted areas.}
    \label{fig:two-images}
\end{figure}

During inference, we utilized a fine-tuned Stable Diffusion model for inpainting on real-world images (see Fig.~\ref{fig:two-images}). A binary image mask specified the region for synthesizing a new object, such as a plant or weed, while a concise text prompt (e.g., \textit{a photo of HoPla Fallopia}) defined the target weed type. Initially, random binary masks were generated dynamically; however, this occasionally caused inpainted regions to overlap with existing objects. To mitigate this, we integrated a fine-tuned object detector (YOLO11X, pre-trained on the COCO dataset~\cite{DBLP:journals/corr/LinMBHPRDZ14}) to exclude predicted regions of interest (ROIs) during mask generation. The same detector was later employed for automatic annotation of the synthetic inpainted images. During image generation, we initially classified weeds into four species: \textit{Cirsium, Convolvulus, Fallopia}, and \textit{Echinochloa}. However, for labeling and weed detection, we reclassified them into two broader botanical categories—dicotyledons (\textit{Cirsium, Convolvulus, Fallopia}) and monocotyledons (\textit{Echinochloa})—to align with herbicide targeting strategies, which focus on botanical groups rather than individual species. Image generation parameters are detailed in Table~\ref{tab:prop}. We employed the Euler Ancestral Discrete scheduler to optimize the trade-off between image quality and computational efficiency. The inference process was configured with 150 steps for optimal image fidelity, and a \textit{strength} parameter of \(0.5\) to control noise, balancing quality, and generation speed. The output resolution was standardized to \(768 \times 512\) pixels to maintain consistency with input dimensions. Additionally, we explored post-training quantization using FP16 and INT8 to reduce memory overhead and accelerate inference.


\begin{table}[h!]
\scriptsize
\renewcommand{\arraystretch}{0.85}

    \centering
    \caption{Properties of inference stage of the fine-tuned Stable Diffusion Model. }
    \begin{tabular}{@{}ll@{}}
    \toprule
        \textbf{Properties} & \textbf{Value} \\ \hline
         Scheduler & Euler Ancestral Discrete\\
         Inference steps & 150 \\
       Guidance scale  & 16 \\
    Strength & 0.5\\
      Image size  & $width: 768 \times height: 512$ \\
      Post-training quantization & FP16, INT8 \\
       
       \bottomrule
    \end{tabular}
    
    \label{tab:prop}
\end{table}

\paragraph{\textbf{Downstream Model Training}}
\label{sub:eva}
Our approach follows a three-stage pipeline: dataset augmentation using quantized Stable Diffusion models, fine-tuning of state-of-the-art object detectors, and post-training quantization with deployment on an edge computing device (cf. Fig.~\ref{fig:quant}).  We begin by quantizing the Stable Diffusion model into three precision formats: FP32, FP16, and INT8. These models generate synthetic images, expanding the original training dataset in controlled increments from 10\% to 200\% with a 10\% increment of its initial size (\(n\)). Two dataset configurations are considered: one using only the original images (\(n\)), and another incorporating synthetic augmentation at various levels.  The original and augmented datasets are then used to fine-tune two state-of-the-art object detection models, YOLO11(l) and RT-DETR(l), both pretrained on the COCO dataset. Each dataset configuration is trained separately to analyze the impact of data augmentation on model performance. 
The training was carried out with 300 epoch, with early stopping to reduce overfitting, with a patient of 30. We use the learning rates 0.01 and 0.001 for the YOLO11(l) and RT-DETR(l) models respectively, with a cosine learning rate schedule for a dynamic adjustment for smooth training. Besides, the online augmentation technique was turned off to maintain training consistency and avoid bias.
 
\paragraph{\textbf{Deployment}} After fine-tuning, the trained models undergo post-training quantization into different bit formats to evaluate the effects of reduced precision. The produced FP32 Torch models coming from downstream detection tasks were converted into \textit{TensorRT} through FP32, FP16, and INT8 quantizations. Prior to deployment, the NVIDIA Jetson Orin Nano was configured to run exclusively the target application by eliminating interfering background processes. Specifically, MAX Power Mode was enabled to ensure that all CPU and GPU cores are active, and the system clocks were set to their maximum frequency. The quantized \textit{TensorRT} models were then deployed on this optimized, resource-constrained edge computing device to assess the real-time capability of the downstream detection models. All models were tested for unbiased evaluation using a fixed validation and test dataset composed exclusively of real-world images.

\section{Results}
\label{sec:res}

We begin by evaluating the impact of various quantization techniques on Stable Diffusion models. To accomplish this, we measured two key metrics: inference time, which indicates the duration taken by the model to generate output, and peak memory usage, which represents the maximum memory consumed during the inference process. The results of our evaluation are summarized in Table~\ref{tab:quantization_comparison}.

\begin{table}[h!] 
\scriptsize
\renewcommand{\arraystretch}{0.85}
    \centering 
    \caption{Performance comparison of different quantization techniques of the Stable Diffusion Model. The best results are highlighted in bold. }
    \begin{tabular}{lcc} 
        \hline 
        \textbf{Quantization} & \shortstack{\textbf{Inference time} \\ ($mean \pm SD$)} & \shortstack{\textbf{Peak memory} \\ ($mean \pm SD$)} \\ 
        \hline 
        FP32 & $16.55 \pm 0.02$ s & $6829.98 \pm 0.005$ MB \\ 
        FP16 & $4.50 \pm 0.033$ s & $3683.45 \pm 0.005$ MB \\ 
       INT8 & $\bm{4.4} \pm \bm{0.05}$ s & $\bm{3683.42} \pm \bm{0.003}$ MB \\ 
        \hline 
    \end{tabular} 
    \label{tab:quantization_comparison} 
\end{table}

The results indicate that the quantization of \textit{FP16} achieves a substantial reduction in inference time, decreasing latency by approximately 72.8\% compared to \textit{FP32} and lowering memory consumption by 46.1\%. \textit{INT8} quantization results in a 73.4\% reduction in inference time compared to \textit{FP32}, while maintaining nearly identical memory efficiency to \textit{FP16}. However, \textit{INT8} exhibits slightly higher variability in inference time, as indicated by its larger standard deviation.

 We evaluated the effect of quantized inpainting augmentation on two downstream models, \textit{YOLO11(l)} and \textit{RT-DETR(l)}, across three quantization settings (\textit{FP32}, \textit{FP16}, and \textit{INT8}). Model performance was assessed using the mAP50 metric.The mean and standard deviation (SD) of mAP50 values across augmentation conditions were computed for each inpainting precision setting (FP32, FP16, INT8) across all quantized downstream models, excluding the `No Augmentation' baseline. A Friedman test confirmed significant differences among precision settings ($p < 0.05$). Post-hoc Wilcoxon signed-rank tests with Bonferroni correction identified pairwise differences. Statistical grouping labels (A, B, C, etc.) indicate settings that are not significantly different (same letter) or significantly different (different letters) (see Tab.~\ref{tab:YOLO_mAP} and~\ref{tab:RT_mAP}). In \textit{YOLO11(l)}, employing high-precision inpainting techniques (FP32 and FP16) resulted in the YOLO11(l) FP32 and FP16 models attaining their peak mAP50 scores of 0.932 with a 200\% augmentation, which marks a modest 0.54\% enhancement compared to the baseline without augmentation. However, the YOLO11(l) INT8 model exhibited a much larger improvement of 6.64\% (from 0.798 to 0.851), indicating that high-precision inpainting helped mitigate the accuracy loss caused by lower model precision. A similar pattern was observed with FP16 inpainting, where the YOLO INT8 model improved by 5.04\% (from 0.814 to 0.855), reinforcing that augmentation can help compensate for lower-precision models.  With INT8 inpainting, the YOLO11(l) INT8 model achieved its peak mAP50 score of 0.862 at 120\% augmentation, reflecting an 8.02\% increase over the no-augmentation baseline (0.798). Interestingly, the YOLO11(l) FP32 and FP16 models reached their highest scores at 200\% augmentation (0.935 mAP50), representing a 1.63\% improvement over baseline. Statistical tests on quantized YOLO11(l) variants revealed significant differences across inpainting precision settings (Friedman test, \( p < 0.05 \)). Post-hoc Wilcoxon tests indicated no significant difference between FP32 and FP16 (\( p > 0.05 \)), while INT8 exhibited significantly lower performance (\( p < 0.05 \)). Consequently, FP32 and FP16 were grouped together (A), with INT8 in a separate group (B). Inpainting precision did not affect YOLO11(l) performance, as these groupings remained consistent across all settings. The\textit{ RT-DETR(l)} model exhibited greater sensitivity to inpainting precision settings. With FP32 inpainting, the RT-DETR(l) FP32 and FP16 models achieved their highest mAP50 scores at 40\% and 90\% augmentation (0.915 and 0.916, respectively), representing an improvement of approximately 1.55\%. The RT-DETR(l) INT8 model showed a slightly higher gain of 5.20\% (from 0.795 to 0.837) at 100\% augmentation, suggesting that moderate augmentation is beneficial. When inpainting was performed at FP16 precision, performance gains were more pronounced, with the RT-DETR(l) FP32 model improving by 3.87\% (from 0.879 to 0.913) and the RT-DETR(l) FP16 model improving by 3.75\% (from 0.880 to 0.913). These results indicate that RT-DETR(l) benefits from FP16 inpainting but does not require as much augmentation as YOLO11(l). Moreover, the RT-DETR(l) INT8 model showed an 9\% improvement (from 0.762 to 0.832) at 140\% augmentation, suggesting that increased augmentation can notably enhance performance in lower-precision settings. With INT8 inpainting, both the RT-DETR(l) FP32 and FP16 models reached their highest performance of 0.917 at 170\% augmentation, demonstrating the effectiveness of augmentation in stabilizing performance under lower-precision inpainting. However, the RT-DETR(l) INT8 model exhibited variability despite achieving an 6.64\% improvement at 130\% augmentation (from 0.782 to 0.834). 
In statistical tests, RT-DETR(l) shows a pattern similar to YOLO11(l), with quantized variants divided into distinct groups: FP32 and FP16 in group (A) and INT8 in group (B), regardless of inpainting precision settings. Similar to YOLO11(l), variations in inpainting precision did not significantly affect RT-DETR(l) performance, maintaining consistent statistical groupings.


\begin{table}[h!]
\footnotesize
\renewcommand{\arraystretch}{0.85}
    \centering
    \caption{Performance of the quantized YOLO11(l) model across different inpainting precision settings and augmentation levels, ranging from no augmentation to 200\%, incorporating both original (Or.) and synthetic (Syn.) data. Results are reported in terms of \textbf{mAP50}, with the highest scores highlighted in \textbf{bold}. Statistical grouping and Mean $\pm$ SD are included to evaluate performance differences and variability across quantization settings.}
    \label{tab:YOLO_mAP}

    \resizebox{\textwidth}{!}{  
        \begin{tabular}{lccccccccc}
            \toprule
            & \multicolumn{3}{c}{\textbf{Inpainting FP32}} 
            & \multicolumn{3}{c}{\textbf{Inpainting FP16}} 
            & \multicolumn{3}{c}{\textbf{Inpainting INT8}} \\
            \cmidrule(lr){2-4} \cmidrule(lr){5-7} \cmidrule(lr){8-10}

            \textbf{Augmentation} 
            & FP32
            & FP16
            & INT8
            & FP32
            & FP16
            & INT8
            & FP32
            & FP16
            & INT8\\

            \midrule
No Augmentation & 0.927 & 0.926 & 0.798 & 0.927 & 0.927 & 0.814 & 0.920 & 0.920 & 0.818 \\
Or. + Syn. (10\%) & 0.924 & 0.924 & 0.808 & 0.921 & 0.922 & 0.825 & 0.928 & 0.928 & 0.821 \\
Or. + Syn. (20\%) & 0.919 & 0.919 & 0.819 & 0.922 & 0.922 & 0.833 & 0.923 & 0.923 & 0.826 \\
Or. + Syn. (30\%) & 0.923 & 0.923 & 0.815 & 0.927 & 0.927 & 0.836 & 0.923 & 0.923 & 0.828 \\
Or. + Syn. (40\%) & 0.927 & 0.927 & 0.828 & 0.921 & 0.921 & 0.800 & 0.926 & 0.927 & 0.834 \\
Or. + Syn. (50\%) & 0.931 & 0.931 & 0.851 & 0.927 & 0.927 & 0.840 & 0.926 & 0.926 & 0.853 \\
Or. + Syn. (60\%) & 0.929 & 0.929 & 0.848 & 0.930 & 0.930 & 0.841 & 0.929 & 0.929 & 0.831 \\
Or. + Syn. (70\%) & 0.901 & 0.901 & 0.822 & 0.929 & 0.928 & 0.845 & 0.893 & 0.893 & 0.798 \\
Or. + Syn. (80\%) & 0.924 & 0.924 & 0.843 & 0.899 & 0.899 & 0.810 & 0.930 & 0.929 & 0.855 \\
Or. + Syn. (90\%) & 0.921 & 0.921 & 0.832 & 0.929 & 0.929 & 0.849 & 0.929 & 0.929 & 0.847 \\
Or. + Syn. (100\%) & 0.924 & 0.923 & 0.832 & 0.925 & 0.925 & 0.850 & 0.930 & 0.930 & 0.845 \\
Or. + Syn. (110\%) & 0.927 & 0.928 & 0.849 & 0.929 & 0.928 & 0.822 & 0.931 & 0.931 & 0.857 \\
Or. + Syn. (120\%) & 0.930 & 0.930 & 0.849 & 0.929 & 0.929 & 0.847 & 0.930 & 0.930 & \textbf{0.862} \\
Or. + Syn. (130\%) & 0.926 & 0.926 & 0.834 & 0.927 & 0.927 & 0.850 & 0.927 & 0.926 & 0.832 \\
Or. + Syn. (140\%) & 0.920 & 0.920 & 0.841 & 0.910 & 0.910 & 0.835 & 0.925 & 0.925 & 0.848 \\
Or. + Syn. (150\%) & 0.927 & 0.927 & \textbf{0.851} & \textbf{0.932} & \textbf{0.931} & \textbf{0.855} & 0.931 & 0.931 & 0.858 \\
Or. + Syn. (160\%) & 0.925 & 0.925 & 0.842 & 0.930 & 0.930 & 0.827 & 0.922 & 0.922 & 0.825 \\
Or. + Syn. (170\%) & 0.931 & 0.931 & 0.840 & 0.929 & 0.929 & 0.843 & 0.929 & 0.928 & 0.858 \\
Or. + Syn. (180\%) & 0.927 & 0.927 & 0.845 & 0.906 & 0.906 & 0.823 & 0.930 & 0.931 & 0.856 \\
Or. + Syn. (190\%) & 0.915 & 0.915 & 0.816 & 0.928 & 0.928 & 0.850 & 0.929 & 0.929 & 0.847 \\
Or. + Syn. (200\%) & \textbf{0.932} & \textbf{0.932} & \textbf{0.851} & 0.909 & 0.909 & 0.820 & \textbf{0.935} & \textbf{0.935} & 0.859 \\
\midrule
$ \text{Mean} \pm \text{SD} \ (\text{Augmentation}) $
 & 0.924 ± 0.007 & 0.924 ± 0.007 & 0.837 ± 0.015 & 0.923 ± 0.009 & 0.923 ± 0.009 & 0.835 ± 0.015 & 0.926 ± 0.008 & 0.926 ± 0.008 & 0.842 ± 0.017 \\
\text{Stat. Group(Augmentation)} &A & A & B & A & A & B & A & A & B \\
\bottomrule
\end{tabular}
}
\end{table}


\begin{table}[h!]
\renewcommand{\arraystretch}{0.85}
    \centering
    \caption{Performance of the quantized RT-DETR(l) model across different inpainting precision settings and augmentation levels, ranging from no augmentation to 200\%, incorporating both original (Or.) and synthetic (Syn.) data. Results are reported in terms of \textbf{mAP50}, with the highest scores highlighted in \textbf{bold}. Statistical grouping and  Mean $\pm$ SD  are included to evaluate performance differences and variability across quantization settings.}
    \label{tab:RT_mAP}

    \resizebox{\textwidth}{!}{  
        \begin{tabular}{lccccccccc}
            \toprule
            & \multicolumn{3}{c}{\textbf{Inpainting FP32}} 
            & \multicolumn{3}{c}{\textbf{Inpainting FP16}} 
            & \multicolumn{3}{c}{\textbf{Inpainting INT8}} \\
            \cmidrule(lr){2-4} \cmidrule(lr){5-7} \cmidrule(lr){8-10}

           \textbf{Augmentation} 
            & FP32
            & FP16
            & INT8
            & FP32
            & FP16
            & INT8
            & FP32
            & FP16
            & INT8\\

            \midrule
No Augmentation & 0.901 & 0.902 & 0.795 & 0.879 & 0.88 & 0.762 & 0.908 & 0.908 & 0.782 \\
        Or. + Syn. (10\%) & 0.896 & 0.896 & 0.789 & 0.904 & 0.905 & 0.815 & 0.883 & 0.884 & 0.758 \\
        Or. + Syn. (20\%) & 0.896 & 0.897 & 0.802 & 0.901 & 0.9 & 0.781 & 0.893 & 0.893 & 0.78 \\
        Or. + Syn. (30\%) & 0.909 & 0.909 & 0.808 & 0.906 & 0.907 & 0.808 & 0.903 & 0.904 & 0.824 \\
        Or. + Syn. (40\%) & \textbf{0.915} & 0.915 & 0.819 & 0.9 & 0.901 & 0.785 & 0.904 & 0.904 & 0.749 \\
        Or. + Syn. (50\%) & 0.909 & 0.909 & 0.722 & 0.909 & 0.909 & 0.753 & 0.882 & 0.882 & 0.765 \\
        Or. + Syn. (60\%) & 0.911 & 0.911 & 0.785 & 0.9 & 0.901 & 0.82 & 0.908 & 0.909 & 0.81 \\
        Or. + Syn. (70\%) & 0.898 & 0.899 & 0.739 & 0.901 & 0.899 & 0.819 & 0.911 & 0.911 & 0.833 \\
        Or. + Syn. (80\%) & 0.904 & 0.904 & 0.798 & 0.911 & 0.91 & 0.811 & 0.909 & 0.909 & 0.768 \\
        Or. + Syn. (90\%) & \textbf{0.915} & \textbf{0.916} & 0.829 & 0.908 & 0.908 & 0.82 & 0.906 & 0.905 & 0.737 \\
        Or. + Syn. (100\%) & 0.907 & 0.907 & \textbf{0.837} & 0.909 & 0.909 & 0.781 & 0.878 & 0.878 & 0.772 \\
        Or. + Syn. (110\%) & 0.901 & 0.902 & 0.779 & 0.91 & 0.911 & 0.818 & 0.91 & 0.91 & 0.793 \\
        Or. + Syn. (120\%) & 0.911 & 0.912 & 0.827 & \textbf{0.913} & \textbf{0.913} & 0.781 & 0.906 & 0.906 & 0.757 \\
        Or. + Syn. (130\%) & 0.873 & 0.874 & 0.767 & 0.907 & 0.907 & 0.77 & 0.906 & 0.907 & \textbf{0.834} \\
        Or. + Syn. (140\%) & 0.901 & 0.901 & 0.82 & 0.901 & 0.901 & \textbf{0.832} & 0.903 & 0.904 & 0.832 \\
        Or. + Syn. (150\%) & 0.906 & 0.906 & 0.817 & 0.905 & 0.905 & 0.817 & 0.903 & 0.903 & 0.83 \\
        Or. + Syn. (160\%) & 0.906 & 0.906 & 0.798 & 0.91 & 0.911 & 0.766 & 0.835 & 0.835 & 0.759 \\
        Or. + Syn. (170\%) & 0.913 & 0.914 & 0.834 & 0.91 & 0.911 & 0.797 & \textbf{0.917} & \textbf{0.917} & 0.794 \\
        Or. + Syn. (180\%) & 0.906 & 0.906 & 0.826 & 0.908 & 0.908 & 0.744 & 0.908 & 0.908 & 0.821 \\
        Or. + Syn. (190\%) & 0.907 & 0.908 & 0.758 & 0.904 & 0.904 & 0.824 & 0.903 & 0.903 & 0.782 \\
        Or. + Syn. (200\%) & 0.900 & 0.900 & 0.794 & 0.902 & 0.902 & 0.805 & 0.903 & 0.903 & 0.796 \\
        \midrule
$ \text{Mean} \pm \text{SD} \ (\text{Augmentation}) $ &0.904 \(\pm\) 0.009 & 0.905 \(\pm\) 0.009 & 0.797 \(\pm\) 0.032 & 0.906 \(\pm\) 0.004 & 0.906 \(\pm\) 0.004 & 0.797 \(\pm\) 0.026 & 0.899 \(\pm\) 0.018 & 0.899 \(\pm\) 0.018 & 0.790 \(\pm\) 0.032 \\
\text{Stat. Group (Augmentation)} &A & A & B & A & A & B & A & A & B \\

\bottomrule

\end{tabular}
}
\end{table}

Furthermore, We evaluate the post-quantization model size and latency on NVIDIA Jetson Orin Nano device intending to deploy in the real-world environment (see Table~\ref{tab:inf}). The inference time and model size results for YOLO and RT-DETR across FP32, FP16, and INT8 precisions (trained on no augmentation and augmentation data)are summarized in Table \ref{tab:inference_time_model_size}. YOLO11(l) consistently outperforms RT-DETR(l) in inference time, with YOLO11(l)-INT8 achieving the fastest execution at $21.75 \pm 1.17$ ms, compared to RT-DETR(l)-INT8 at $45.13 \pm 15.12$ ms. A similar trend is observed in FP32 and FP16, where YOLO11(l)-FP32 and YOLO-FP16 achieve $63.87 \pm 4.01$ ms and $32.48 \pm 2.68$ ms, respectively, while RT-DETR(l)-DETR-FP32 and RT-DETR(l)-FP16 remain slower at $88.06 \pm 5.40$ ms and $51.55 \pm 1.97$ ms. In terms of model size, INT8 compression achieves the highest reduction, where YOLO11(l)-INT8 (30.8 MB) is smaller and faster than RT-DETR(l)-INT8 (47.4 MB).

\begin{table}[h!]

\scriptsize
    \centering
    \renewcommand{\arraystretch}{0.85}
    \caption{Inference Time $(mean \pm standard\ deviation (SD))$ \& model size comparison of YOLO11(l) and RT-DETR(l) trained on both non-augmented and augmented data, evaluated on NVIDIA Jetson Orin Nano.}
    \label{tab:inf}
    \label{tab:inference_time_model_size}
    \begin{tabular}{lccc ccc}
        \toprule
        \multirow{2}{*}{\textbf{Model}} & \multicolumn{3}{c}{\textbf{Inference Time $(Mean \pm SD) (ms)$}} & \multicolumn{3}{c}{\textbf{Model Size $(MB)$}} \\
        \cmidrule(lr){2-4} \cmidrule(lr){5-7}
        & \textbf{FP32} & \textbf{FP16} & \textbf{INT8} & \textbf{FP32} & \textbf{FP16} & \textbf{INT8} \\
        \midrule
        \textbf{YOLO11(l)}    & 63.87 ± 4.01  & 32.48 ± 2.68  & 21.75 ± 1.17  & 98.8  & 51.7  & 30.8  \\
        \textbf{RT-DETR11(l)} & 88.06 ± 5.40  & 51.55 ± 1.97  & 45.13 ± 15.12 & 125.0 & 64.6  & 47.4  \\
        \bottomrule
    \end{tabular}
\end{table}
\section{Discussion}
\label{sec:Dis}

The reported results highlight the impact of quantization on the computational efficiency and performance of the Stable Diffusion model and its downstream models, including YOLO11(l) and RT-DETR(l). In the Stable Diffusion Model, quantization techniques such as FP16 and INT8 substantially reduce inference time and peak memory usage in comparison to FP32(cf.~\ref{tab:quantization_comparison}). However, the difference in peak memory usage and inference time between FP16 and INT8 in the Stable Diffusion model is minimal. Several factors contribute to this observation. First, the model's complex operations, such as attention mechanisms and residual connections, do not fully exploit the benefits of INT8 quantization. Additionally, INT8 quantization introduces computational overhead due to frequent dequantization and re-quantization steps, which counteract its potential performance improvements~\cite{dettmers2022llmINT88bitmatrixmultiplication}. Furthermore, modern hardware optimizations are often designed to favor FP16, as many GPUs and NPUs are better optimized for mixed-precision computations, which might lead to almost similar performance between FP16 and INT8 ~\cite{zhou2025etbench}. Furthermore, this study indicates that inpainting augmentation can help improve performance lost due to quantization. The effectiveness of this approach depends on the inpainting precision (FP32, FP16, INT8) and the model architecture used.  On the downstream models, high-precision settings(FP32 and FP16) provide slight improvements. In contrast, in the INT8 setting, the accuracy is more affected by quantization degradation, which shows a much greater recovery in performance when using synthetic data in all quantized variations of the Stable Diffusion Model. Moreover, our findings further reveal that the benefits of inpainting augmentation are architecture-specific. The YOLO11(l) model appears to leverage synthetic augmentation more effectively, especially in its INT8 configuration, whereas RT-DETR(l) demonstrates more gradual and sometimes variable improvements. However, the relationship between the augmentation and the detection accuracy is varied. This could be a reason of the automated annotation of without further checking, which aligns with the findings of~\cite{modak2024generative}.Nevertheless, this experiment was conducted on a single training run using a randomly selected subset of synthetic images for each augmentation combination. To improve statistical reliability and reduce bias, future experiments will incorporate stratified subsampling across at least 10 independent sets for more robust evaluation. Besides, the practical advantages of quantization and inpainting augmentation are evident in the reductions in model size and inference latency—particularly for the INT8 configuration. These efficiency gains are critical for deploying models in resource-constrained environments, where minimizing memory and computational requirements is essential. By effectively combining the benefits of quantization and synthetic inpainting, we can achieve high-performance models that are well-suited for real-time applications.

\section{Conclusion}
\label{sec:con}

This study explores the potential of quantizing the Stable Diffusion Model and downstream models (YOLO11(l) and RT-DETR(l)) integrated into an GenAI-based weed detection pipeline as introduced in~\cite{10.1007/978-3-031-66146-4_8}. Post-training quantization of the incorporated Stable Diffusion Model to FP16 and INT8 can substantially reduce latency and computational cost without degrading downstream performance during augmentation.
Additionally, quantization improves the inference speed of downstream models, while synthetic image augmentation has been observed to be able to mitigate the performance loss associated with INT8 quantization in YOLO11(l) and RT-DETR(l). Our findings also suggest that the effectiveness of inpainting augmentation varies when looking at different models. Moreover, performance variability appears to be linked to the quality of automated annotations, suggesting that refined annotation strategies could further enhance results. Future research will investigate other quantization strategies of the Stable Diffusion Model, such as BF16, FP8, and FP4 to reduce latency on resource-constrained devices, enabling deployment on platforms such as the NVIDIA Jetson Orin Nano or NPUs. If augmentation pipeline latency is reduced, our method could be integrated into intelligent systems architectures, such as the MLOC~\cite{stein2025organic} from Organic Computing, for applications in intelligent agricultural robots. These architectures include reflection layers that monitor adaptation layers controlling the system (SuOC). When performance drops are detected, reflection layers trigger reconfigurations, such as triggering continual learning to compensate for lacking robustness in corner cases or new environmental conditions~\cite{stein2021reflective}, to restore performance by addressing knowledge gaps~\cite{Stein2018} through on-demand synthetic data training.

\subsubsection*{Acknowledgements.}
\label{subsec:ack}
This research was conducted within the scope of the project ``Hochleistungssensorik für smarte Pflanzenschutzbehandlung (HoPla)" (grant no. 13N16327), supported by the Federal Ministry of Education and Research (BMBF) and VDI Technology Center based on a decision by the German Bundestag.

%

\bibliographystyle{splncs04}
\bibliography{bibliography}

\end{document}